\documentclass[11pt]{article}

\usepackage[]{acl}
    \usepackage{array}
    \usepackage{ragged2e}
    \usepackage{pifont}        
    \usepackage{xcolor}
    \usepackage{soul}  
\usepackage{times}
\usepackage{latexsym}
\usepackage{booktabs}
\usepackage{amssymb} 
\usepackage{url}
\usepackage[table]{xcolor}
\usepackage[T1]{fontenc}

\usepackage[utf8]{inputenc}

\usepackage{microtype}
\usepackage{multirow}
\usepackage{colortbl}
\usepackage{xcolor}
\usepackage{booktabs}
\usepackage[table]{xcolor}
\usepackage{fontawesome5}

\usepackage{inconsolata}

\usepackage{graphicx} 
\usepackage{amsmath}
\usepackage{tikz}
\usepackage{adjustbox}
\usepackage{pgffor}

\usetikzlibrary{arrows.meta,positioning,calc}

\newcommand{\smallstdev}[1]{\scalebox{0.75}{$\pm$#1}}%

\definecolor{darkblue}{rgb}{0, 0, 0.5}
\hypersetup{colorlinks=true, citecolor=darkblue, linkcolor=darkblue, urlcolor=darkblue}

\title{Improving Argument Saliency Coverage in Small LLMs for Long Legal Opinion Summarization via Sequence-Level Distillation}

\author{Mohamed Elaraby\textsuperscript{1}, \quad
  Ahmed Elhady\textsuperscript{2}, \quad
  Diane Litman\textsuperscript{1} \\
  \textsuperscript{1}University of Pittsburgh, Pittsburgh, PA, USA \\
  \textsuperscript{2}HiTZ Center, University of the Basque Country (UPV/EHU) \\
  \texttt{\{mse30,dlitman\}@pitt.edu}, \quad \texttt{ahmed.salemmohamed@ehu.eus} \\}
  
\begin{document}
\maketitle
\begin{abstract}
We show that sequence-level distillation from a capable long-context teacher model is a simple, annotation-free, and data-efficient strategy for improving argument saliency coverage in long legal opinion summarization, where small LLMs often struggle to retain the most salient argumentative content. Across student model sizes, distillation consistently surpasses tuning on expert-written summaries in our legal-opinion setting. We further demonstrate that most gains are achieved with as few as $\sim$10 training summaries, highlighting the strong data efficiency of teacher-generated supervision. Finally, we find that summary distillation is sufficient for improvements: reasoning-chain distillation remains competitive with summary-only distillation, but provides marginal benefit when combined with summary supervision.
\end{abstract}

\section{Introduction}

Summarizing long legal opinions requires preserving salient argument roles that are sparse across long inputs yet critical to include in the final summary~\cite{xu2021toward, elaraby2022arglegalsumm}. Recent work shows that LLMs still struggle to prioritize the same salient information as humans in summarization tasks~\cite{trienes2025behavioral}, a challenge that becomes especially apparent in legal opinion summarization, where important argumentative content is dispersed across thousands of tokens~\cite{elaraby-litman-2026-arc}. This problem is further amplified in smaller models, which limits the development of efficient and reliable summarization systems for high-stakes legal settings.

\begin{figure}[t]
    \centering
    \includegraphics[width=.95\columnwidth]{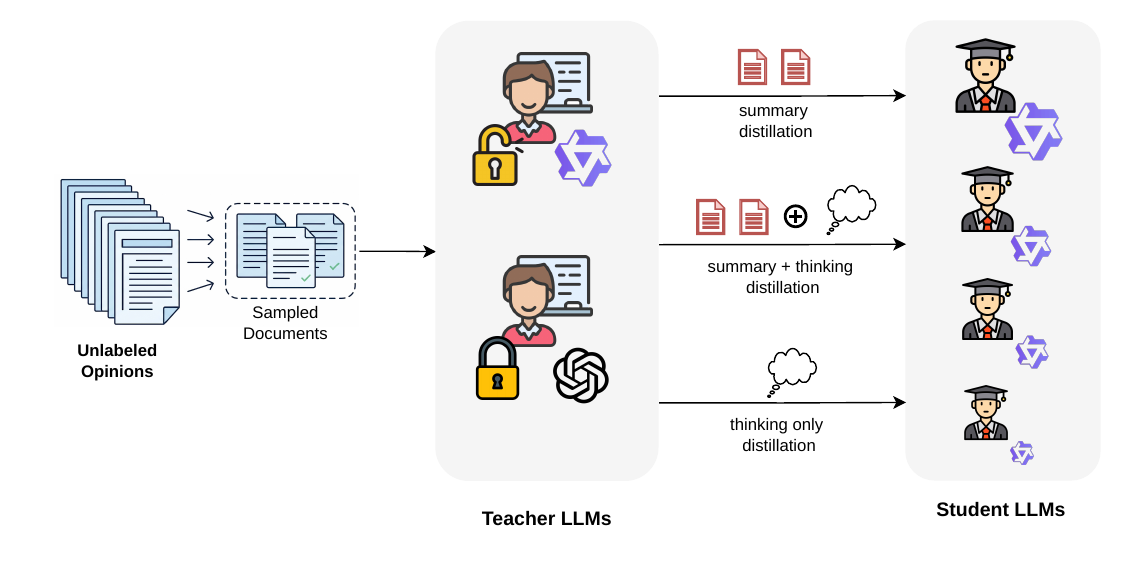}
    \caption{Distillation strategies from proprietary (\faLock) and open-weight (\faLockOpen) teacher models to student models of various sizes.}
    \label{fig:distillation_techniques}
\end{figure}

While tuning models on expert-written summaries is a plausible solution, obtaining domain-expert annotations remains costly and difficult to scale. Sequence-level distillation~\cite{kim-rush-2016-sequence}, which transfers generation behavior from a stronger teacher through teacher-generated sequences, offers an annotation-free alternative. Recent work has explored related forms of summarization distillation by training smaller models on LLM-generated summaries~\cite{liu-etal-2024-learning,zhu2025factual} or prompted structured rationales~\cite{jiang-etal-2024-trisum, wang-etal-2025-distilling}. However, prior work has largely focused on news-style benchmarks, such as \texttt{CNN/DailyMail}~\cite{hermann2015teaching} and \texttt{XSum}~\cite{narayan-etal-2018-dont}, emphasizing faithfulness, fluency, or generic summary quality rather than saliency coverage in long, high-stakes domains. We therefore ask: \textit{To what extent can sequence-level distillation improve argument saliency coverage in long legal opinion summarization?} We empirically study this question using small long-context LLMs ($<10$B parameters) from the \texttt{Qwen3} family. Motivated by recent advances in reasoning-capable LLMs, we examine whether \textit{native reasoning traces} provide useful supervision beyond teacher-generated summaries. While intermediate reasoning supervision has been shown to improve smaller models on reasoning-intensive tasks~\cite{guo-etal-2025-deepseek-r1}, its value for long-context summarization remains unclear.

We compare three supervision strategies (Figure~\ref{fig:distillation_techniques}): \textit{(1) summary-only}, trained on teacher-generated summaries; \textit{(2) reasoning-trace-only}, trained only on the teacher's native reasoning trace; and \textit{(3) reasoning-trace + summary}, trained on both signals. We evaluate proprietary and open-weight teachers, comparing distillation against expert-summary tuning and structured inference-time planning baselines. Finally, we vary the number of distillation examples to study data efficiency, a question largely unexplored in prior summarization distillation work, which typically relies on thousands of generated summaries.

Our findings are twofold. First, \textit{sequence-level distillation is an effective annotation-free strategy for improving argument saliency coverage in long legal summarization}: it consistently surpasses both tuning on expert-written summaries and structured inference-time planning baselines. Second, \textit{distillation is simple and data-efficient}: most gains emerge with as few as $\sim10$ training documents, and teacher-generated summaries alone are sufficient, with reasoning-trace supervision providing only marginal additional benefit. 

\section{Related Work}

\noindent \textbf{Legal Opinion Summarization and Saliency Coverage.}
While prior work on LLM-based legal opinion summarization has identified
faithfulness as a persistent issue~\cite{heddaya-etal-2025-casesumm,
zhong-litman-2025-discourse}, recent work shows that coverage gaps of
salient argumentative content can be even larger~\cite{elaraby-litman-2026-arc}. Existing approaches to legal coverage rely on external argument labeling~\cite{elaraby2022arglegalsumm} or reranking
pipelines~\cite{elaraby-etal-2023-towards}; in news, coverage is improved through structured planning over entities~\cite{adams-etal-2023-sparse} or events~\cite{gantt-etal-2024-event, walden-etal-2025-cross}, which we adapt as a baseline.
\textit{Unlike the news domain, which often involves short-context, entity-focused settings, we study
sequence-level distillation for argument saliency coverage in long legal opinions,
where salient content is sparse across thousands of tokens.}

\noindent \textbf{Distillation from LLMs for Summarization.}
Sequence distillation has long been used to transfer generation behavior from larger models to smaller ones~\citep{kim-rush-2016-sequence}. Recent work applies this idea to summarization by training smaller models on LLM-generated summaries~\citep{liu-etal-2024-learning}, as well as using richer intermediate supervision signals such as structured rationales~\citep{jiang-etal-2024-trisum,wang-etal-2025-distilling}. However, prior work has largely focused on general-domain or news-style benchmarks, emphasizing fluency, faithfulness, or generic summary quality. Long legal opinions pose a different challenge: salient argumentative content is sparse and dispersed across thousands of input tokens, making content selection and omission errors central to summary quality. \textit{We extend distillation work to empirically study its effect in improving argument saliency in small LLMs.}

\section{Distillation Setup}

Given a prompt $P$ and a legal opinion $L$, a teacher model $M$
generates a native reasoning trace $\tau$ followed by a summary $S$,
denoted $(\tau,S)\sim M(P,L)$.
We empirically study sequence distillation under three supervision settings
differing only in the supervision sequence $\mathcal{G}$.
Student models $m$ are trained using supervised fine-tuning
(SFT)~\cite{ouyang2022training}:

\begin{equation}
\mathcal{L}_{\mathrm{SFT}}
=
-\frac{1}{|\mathcal{G}|}
\sum_{t=1}^{|\mathcal{G}|}
\log P_m(g_t\mid g_{<t},P,L),
\end{equation}

where $\mathcal{G}$ denotes the supervised target sequence. We use these settings to test whether the teacher's native reasoning trace provides supervision beyond the final summary. In particular, while standard rationale-distillation setups often supervise both reasoning and the final answer, our reasoning-only setting optimizes only over $\tau$, masking the final summary from the loss. This tests whether the reasoning trace alone can transfer the teacher's intermediate saliency-selection process, rather than simply teaching the student to imitate summary outputs. This yields three settings:
\textit{(1) summary-only distillation}, $\mathcal{G}=S$;
\textit{(2) reasoning distillation}, $\mathcal{G}=\tau$;
and \textit{(3) reasoning + summary distillation},
$\mathcal{G}=\tau\oplus S$. 

\section{Experimental Setup}
\subsection{Dataset}
We use the Canadian Legal Information Institute (\texttt{CANLII}) dataset, which contains long legal opinions paired with expert annotations of argumentative roles relevant for summarization~\cite{xu2021toward}. Argument roles are annotated using the \textbf{IRC} scheme~\cite{xu2021toward}, consisting of \textbf{Issues} (legal questions raised in a case), \textbf{Reasons} (legal principles and justifications underlying decisions), and \textbf{Conclusions} (final rulings resolving the issues).\footnote{Appendix~\ref{app:canlii_examples} shows examples of argument roles.} The full dataset contains approximately $28$k legal opinion--summary pairs. A subset of $1{,}049$ \texttt{CANLII} opinions is annotated at the sentence level with argumentative roles for both source documents and summaries~\cite{xu2021toward},\footnote{Data obtained under a license agreement with CANLII.} enabling evaluation of argument saliency coverage; we use these as our test split.  For training, we sample $1{,}000$ unannotated legal opinions that do not overlap with the annotated test set, truncating inputs to $4{,}096$ tokens. This is motivated by prior work showing that supervised fine-tuning on shorter contexts can generalize to longer-context settings~\cite{gao-etal-2025-train, grattafiori2024llama}.
Training documents average $2{,}990$ words, compared to $4{,}382$ words in the test split.

\subsection{Models}
We use two teacher models representing complementary settings: the proprietary \texttt{GPT-5-mini} and the open-weight \texttt{Qwen3-14B}. Since proprietary models do not expose reasoning traces, reasoning-trace distillation is applied only with \texttt{Qwen3-14B}. As student models, we use the \texttt{Qwen3} family ranging from 0.6B to 8B parameters, allowing controlled comparison across model sizes under a shared architecture. Qwen3 additionally supports reasoning-enabled and reasoning-disabled inference\footnote{via the \texttt{/no\_think} prompt suffix.} using the same underlying weights, enabling us to separate the effect of test-time reasoning from distillation itself. For each of the $1{,}000$ training documents, teachers generate summaries of approximately $250$ words, matching the average expert summary length in the test split ($270$ words), using temperature $0.7$ and nucleus sampling ($p=0.95$).\footnote{Training details and prompts are in Appendix~\ref{app:train_summ_prompt}.}

\newcommand{\gc}[1]{\cellcolor{gray!12}#1}

\subsection{Baselines and Evaluation}

We compare distillation against three baselines. The \textbf{zero-shot} (\textbf{B}) baseline generates summaries from each student in thinking and non-thinking modes. The \textbf{expert tuning} (\textbf{E}) baseline fine-tunes each student on expert-written summaries from the same $1{,}000$ documents, enabling direct comparison between teacher-generated and expert supervision.
Motivated by prior structured planning work~\cite{adams-etal-2023-sparse, wang-etal-2025-distilling}, the \textbf{Chain-of-Arguments} (\textbf{CoA}) baseline disables native thinking and prompts the student to generate an argument plan before the final summary,\footnote{Prompt in Appendix~\ref{app:CoA}.} isolating inference-time planning from fine-tuning.
\begin{table}[t]
\small
\centering
\begin{adjustbox}{max width=\columnwidth}
\begin{tabular}{l cccc}
\toprule
\textbf{Setup}
  & \textbf{Qwen3-0.6B}
  & \textbf{Qwen3-1.7B}
  & \textbf{Qwen3-4B}
  & \textbf{Qwen3-8B} \\
\midrule
\multicolumn{5}{c}{\textit{Teacher baseline}} \\
\midrule
\textit{GPT-5-mini}
  & \multicolumn{4}{c}{\textit{.742\smallstdev{.166}}} \\
\textit{Qwen3-14B}
  & \multicolumn{4}{c}{\gc{\textit{.738\smallstdev{.172}}}} \\
\midrule
\multicolumn{5}{c}{\textbf{Baselines}} \\
\midrule
\multirow{2}{*}{\textbf{B} }
  & .408\smallstdev{.219} & .522\smallstdev{.218} & .631\smallstdev{.188} & .658\smallstdev{.184} \\
  & \gc{.514\smallstdev{.222}} & \gc{.530\smallstdev{.209}} & \gc{.635\smallstdev{.187}} & \gc{.654\smallstdev{.186}} \\
\addlinespace
\multirow{2}{*}{\textbf{E}}
  & .599\smallstdev{.176} & .599\smallstdev{.176} & .571\smallstdev{.206} & .526\smallstdev{.235}$^\ddagger$ \\
  & \gc{.600\smallstdev{.175}} & \gc{.618\smallstdev{.210}} & \gc{.613\smallstdev{.197}} & \gc{.510\smallstdev{.229}$^\ddagger$} \\
\addlinespace
\multirow{1}{*}{\textbf{CoA}}
  & .578\smallstdev{.219} & .624\smallstdev{.203} & .679\smallstdev{.171} & .698\smallstdev{.169} \\
\midrule
\multicolumn{5}{c}{\textbf{Distillation (GPT-5-mini Teacher)}} \\
\midrule
\multirow{2}{*}{\textbf{S}}
  & .553$^*$\smallstdev{.213} & .652$^*$\smallstdev{.196} & .712$^*$\smallstdev{.175} & .741$^*$\smallstdev{.168} \\
  & \gc{.554$^*$\smallstdev{.210}} & \gc{.658$^*$\smallstdev{.193}} & \gc{.714$^*$\smallstdev{.172}} & \gc{.735$^*$\smallstdev{.170}} \\ \midrule
\multicolumn{5}{c}{\textbf{Distillation (Qwen3-14B Teacher)}} \\
\midrule
\multirow{2}{*}{\textbf{S}}
  & .621$^*$\smallstdev{.209} & .717$^*$\smallstdev{.177} & .744$^*$\smallstdev{.160} & .781$^*$\smallstdev{.150} \\
  & \gc{.628$^*$\smallstdev{.204}} & \gc{.723$^*$\smallstdev{.177}} & \gc{.741$^*$\smallstdev{.165}} & \gc{.784$^*$\smallstdev{.151}} \\
\addlinespace
\multirow{2}{*}{\textbf{R}}
  & .622$^*$\smallstdev{.200} & .718$^*$\smallstdev{.175} & .742$^*$\smallstdev{.163} & .779$^*$\smallstdev{.154} \\
  & \gc{\textbf{.632}$^*$\smallstdev{.200}} & \gc{.716$^*$\smallstdev{.179}} & \gc{.745$^*$\smallstdev{.165}} & \gc{.774$^*$\smallstdev{.161}} \\
\addlinespace
\multirow{2}{*}{\textbf{SR}}
  & .630$^*$\smallstdev{.210} & .722$^*$\smallstdev{.177} & .751$^*$\smallstdev{.162} & .777$^*$\smallstdev{.154} \\
  & \gc{\textbf{.632}$^*$\smallstdev{.199}} & \gc{\textbf{.733}$^*$\smallstdev{.169}} & \gc{\textbf{.753}$^*$\smallstdev{.161}} & \gc{\textbf{.785}$^*$\smallstdev{.154}} \\
\bottomrule
\end{tabular}
\end{adjustbox}
\caption{
\texttt{ARC}\textsubscript{score} (mean$\pm$std.). 
\textit{Italicized \textbf{Teacher baseline} rows} report each teacher's standalone \texttt{ARC}\textsubscript{score} on the test set (a single value per teacher, spanning the student columns), serving as a reference point for the distilled students below. 
\textbf{B}=zero-shot; \textbf{E}=expert-tuning; 
\textbf{CoA}=Chain-of-Arguments prompting; 
\textbf{S}=summary-only; \textbf{R}=reasoning-only; 
\textbf{SR}=summary+reasoning. 
\colorbox{gray!12}{\,Shaded\,} rows use thinking inference. 
$^*$/$^\ddagger$ denote significantly above / below \textbf{B} 
using Mann--Whitney U test ($p<0.05$). 
\bf{Bold} = best.
}
\label{tab:arc_1000_distillation}
\end{table}

We evaluate argument saliency coverage using \texttt{ARC}\textsubscript{score}~\cite{elaraby-litman-2026-arc}, which compares generated summaries $S$ against the set of expert-annotated argumentative roles $\mathcal{A}$. The metric computes the proportion of arguments in $\mathcal{A}$ supported by $S$ using an LLM-based atomic verifier, and has been shown to achieve the strongest correlation with expert coverage judgments~\cite{elaraby-litman-2026-arc} on the expert-annotated legal summaries of~\citet{elaraby-etal-2024-adding}. Unlike decomposition-based metrics such as \texttt{FactScore}~\cite{min-etal-2023-factscore} or claim-verification approaches such as \texttt{MiniCheck}~\cite{tang-etal-2024-minicheck}, \texttt{ARC}\textsubscript{score} distinguishes between \textit{Missing Errors} (\textbf{ME}), where salient arguments are omitted, and \textit{Factual Errors} (\textbf{FE}), where generated summaries contradict one or more salient arguments, enabling finer-grained analysis of whether gains stem from improved coverage or factual consistency. \footnote{\texttt{ARC}\textsubscript{score} details in Appendix \ref{app:arc_score}}

\section{Results and Discussion}

\subsection{Distillation Findings}

\definecolor{summarybg}{HTML}{F2F2F2}
\definecolor{thoughtbg}{HTML}{FFF2CC}
\definecolor{sumthoughtbg}{HTML}{EADCF8}
\definecolor{expertbg}{HTML}{E2F0D9}
\definecolor{teacherbg}{HTML}{DDEBF7}
\definecolor{groupbg}{HTML}{E8E8E8}

\begin{table}[t]
\small
\centering
\begin{adjustbox}{max width=\linewidth}
\begin{tabular}{l cccc cccc}
\toprule
\textbf{Setup}
  & \multicolumn{2}{c}{\textbf{Qwen3-0.6B}}
  & \multicolumn{2}{c}{\textbf{Qwen3-1.7B}}
  & \multicolumn{2}{c}{\textbf{Qwen3-4B}}
  & \multicolumn{2}{c}{\textbf{Qwen3-8B}} \\
  & \textbf{FE} & \textbf{ME}
  & \textbf{FE} & \textbf{ME}
  & \textbf{FE} & \textbf{ME}
  & \textbf{FE} & \textbf{ME} \\
\midrule

\multicolumn{9}{c}{\textbf{Baselines}} \\
\midrule
\multirow{2}{*}{\textbf{B}}
  & $10.20$ & $55.87$ & $11.82$ & $43.65$ & $12.22$ & $31.76$ & $11.86$ & $30.01$ \\
  & \gc{$10.85$} & \gc{$47.28$} & \gc{$11.84$} & \gc{$42.78$} & \gc{$12.37$} & \gc{$31.49$} & \gc{$12.17$} & \gc{$29.84$} \\
\midrule

\multicolumn{9}{c}{\textbf{Distillation (GPT-5-mini Teacher)}} \\
\midrule
\multirow{2}{*}{\textbf{S}}
  & $14.40$ & $36.31$ & $13.77$ & $28.33$ & $12.15$ & $22.48$ & $10.87$ & $21.12$ \\
  & \gc{$14.22$} & \gc{$36.71$} & \gc{$13.96$} & \gc{$27.20$} & \gc{$11.91$} & \gc{$22.40$} & \gc{$10.89$} & \gc{$20.56$} \\
  \midrule

\multicolumn{9}{c}{\textbf{Distillation (Qwen3-14B Teacher)}} \\
\midrule
\multirow{2}{*}{\textbf{S}}
  & $12.22$ & $31.31$ & $10.89$ & $22.42$ & $10.37$ & $19.78$ & $9.05$ & $17.40$ \\
  & \gc{$12.59$} & \gc{$29.87$} & \gc{$10.82$} & \gc{$22.36$} & \gc{$10.39$} & \gc{$20.24$} & \gc{$8.38$} & \gc{$17.74$} \\
\addlinespace
\multirow{2}{*}{\textbf{R}}
  & $12.02$ & $31.20$ & $10.20$ & $22.78$ & $10.42$ & $19.88$ & $8.28$ & $18.57$ \\
  & \gc{$12.66$} & \gc{$30.69$} & \gc{$10.24$} & \gc{$23.46$} & \gc{$10.48$} & \gc{$19.79$} & \gc{$8.75$} & \gc{$18.68$} \\
\addlinespace
\multirow{2}{*}{\textbf{SR}}
  & $12.23$ & $30.31$ & $11.06$ & $22.36$ & $9.72$ & $19.71$ & $8.31$ & $17.75$ \\
  & \gc{$12.55$} & \gc{$30.67$} & \gc{$10.21$} & \gc{$22.33$} & \gc{$9.59$} & \gc{$19.56$} & \gc{$8.23$} & \gc{$17.53$} \\

\bottomrule
\end{tabular}
\end{adjustbox}
\caption{
  Factual Errors (\textbf{FE}) and Missing Errors (\textbf{ME}) in $\%$ per distillation setup from \texttt{ARC}$_\text{score}$ diagnostic evaluation. Setup abbreviations and row shading follow Table~\ref{tab:arc_1000_distillation}.
}
\label{tab:fe_me_distillation}
\end{table}

Table~\ref{tab:arc_1000_distillation} presents our main results using $1{,}000$ training documents capped at $4{,}096$ tokens.\footnote{Increasing training context length yields only marginal additional gains; see Appendix~\ref{app:len_scale}.} 
\textit{$(1)$ Distillation significantly improves saliency coverage.} Across all student model sizes, distillation yields statistically significant improvements over the zero-shot baseline (\textbf{B}; $p<0.05$). It also consistently outperforms both expert-summary tuning (\textbf{E}) and the Chain-of-Arguments (\textbf{CoA}) inference-time planning baseline, suggesting that sequence-level distillation is more effective for transferring argument saliency behavior than either limited expert supervision or structured prompting alone. \footnote{Paired Wilcoxon signed-rank tests against \textbf{E} and \textbf{CoA} confirm these differences for \texttt{Qwen3-1.7B} and larger; at \texttt{Qwen3-0.6B}, distillation from \texttt{Qwen3-14B} is statistically comparable to expert tuning (full results in Appendix~\ref{app:significance}).} 
\textit{$(2)$ Distillation performance is teacher-dependent.} While both \texttt{GPT-5-mini} and \texttt{Qwen3-14B} improve coverage, \texttt{Qwen3-14B} consistently achieves the strongest downstream results. Notably, this gap does not reflect a difference in teacher summarization quality: on the test set, the two teachers achieve nearly identical standalone \texttt{ARC}\textsubscript{score} (\texttt{GPT-5-mini}: $0.742{\pm}0.166$; \texttt{Qwen3-14B}:$0.738{\pm}0.172$), yet produce substantially different downstream distillation performance. We hypothesize that this partly reflects closer distributional alignment between the open-weight teacher and the \texttt{Qwen3} student family, rather than teacher output quality alone.\footnote{Appendix~\ref{app:perplexity} analyzes teacher and expert-summary perplexity, while Appendix~\ref{app:expert_tuning} covers expert-tuned students and their common failures.} By contrast, \texttt{GPT-5-mini} yields smaller gains and underperforms \textbf{CoA} for the smallest 0.6B student.
\textit{$(3)$ Summary-only supervision is sufficient.} Native-reasoning distillation (\textbf{R}) remains competitive with summary-only distillation (\textbf{S}), while combining reasoning traces and summaries (\textbf{SR}) yields only marginal gains over \textbf{S} ($+0.001$--$+0.011$). This suggests that, for argument saliency coverage, teacher-generated summaries already encode much of the useful supervision signal. This contrasts with reasoning-intensive settings, where intermediate rationale or reasoning supervision can substantially improve smaller models~\cite{hsieh-etal-2023-distilling, guo-etal-2025-deepseek-r1, muennighoff-etal-2025-s1}.

Table~\ref{tab:fe_me_distillation} provides a diagnostic view of how distillation affects different error types. We observe that \textit{Missing Errors} (\textbf{ME}) --- the dominant error type in the zero-shot baseline --- decrease consistently across student sizes, whereas \textit{Factual Errors} (\textbf{FE}) decrease only for larger students ($\geq 4B$). We focus this breakdown on distillation settings, as the central diagnostic question is whether distillation gains stem from improved coverage or improved faithfulness. For example, Qwen3-14B summary distillation reduces ME from $55.87\%$ to $31.31\%$ for the 0.6B student, while FE slightly increases from $10.20\%$ to $12.22\%$. In contrast, the 8B student shows reductions in both ME ($30.01\% \rightarrow 17.40\%$) and FE ($11.86\% \rightarrow 9.05\%$).\footnote{Appendix~\ref{app:relative_reduction} visualizes the relative change in FE and ME.}

\subsection{Effect of Scaling Distillation Samples}
\label{subsec:scale_samples}
\begin{figure}[t]
    \centering
    \includegraphics[width=0.95\columnwidth]{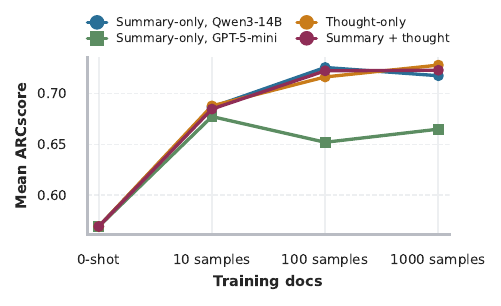}
    \caption{
    Few-shot distillation averaged across students. 
    }
    \label{fig:fewshot_distillation}
\end{figure}

Figure~\ref{fig:fewshot_distillation} shows that
argument coverage improves significantly with as few as $10$ training 
documents across both proprietary and open-weight teachers (Mann-Whitney U 
test, $p<0.05$). Scaling data further yields diminishing returns: 
\texttt{Qwen3-14B} shows only marginal gains beyond $100$ documents and 
near-saturation at $1000$.\footnote{Higher LoRA ranks do not help; see Appendix~\ref{app:lora_impact}.} \textit{GPT-5-mini}, however, follows a less stable trajectory---coverage 
drops at $100$ documents before partially recovering at $1{,}000$---further 
confirming that scaling training data yields diminishing returns regardless 
of the teacher. To ensure these few-shot gains are not an artifact of which 
documents are sampled, we re-ran the $10$- and $100$-document settings with 
a second random seed for \texttt{Qwen3-1.7B} and \texttt{Qwen3-4B}; the 
trend is stable across seeds, differing by at most $0.031$ 
\texttt{ARC}\textsubscript{score}.
\footnote{Detailed breakdowns are in Appendix~\ref{app:few_shot_detailed}.}

\section{Conclusion and Future Work}
We show that sequence-level distillation is a simple, data-efficient strategy for improving argument saliency in long legal summarization, outperforming expert-supervised tuning with as few as $10$ training examples on the \texttt{CANLII} benchmark. Our results demonstrate that, in this setting, summary-only supervision is sufficient, as reasoning-chain distillation provides only marginal gains. We note that these findings are established on a single legal dataset, one primary coverage metric, and the \texttt{Qwen3} student family, and assume access to unlabeled in-domain opinions; establishing their generality across datasets, metrics, and model families is an important direction. Future work will explore alternative distillation objectives and extend this framework to additional long-context domains and evaluation metrics beyond argument saliency. Two further directions follow from our analysis: replacing the human-annotated atomic facts in \texttt{ARC}\textsubscript{score} with LLM-extracted facts, which would relax the dependence on annotated corpora and allow both the evaluation and the distillation pipeline to scale to larger, unannotated legal collections; and a curriculum-style adaptation that distills toward the more on-distribution teacher summaries before a harder second-stage adaptation on expert data, potentially making the expert signal more learnable than direct expert tuning.

\section*{Limitations}

\noindent \textbf{Distillation and Learning Strategy.}
This work focuses on simple sequence-level distillation using standard supervised fine-tuning (SFT), motivated by isolating the effect of distillation itself. While effective, prior work suggests that alternative objectives such as contrastive learning or preference optimization may provide stronger supervision signals than standard SFT. Additionally, all experiments employ parameter-efficient LoRA tuning rather than full-parameter fine-tuning, largely due to GPU constraints. Exploring whether the observed trends persist under stronger optimization strategies remains an important direction for future work.

\noindent \textbf{Limited Domain and Evaluation Scope.}
Our conclusions are established on a single legal-domain dataset (\texttt{CANLII}), one primary evaluation metric (\texttt{ARC}\textsubscript{score}), one student model family (\texttt{Qwen3}), and two teacher models, and should be read as scoped to this setting rather than as general claims about long-document summarization. While argument coverage is particularly important in legal reasoning, coverage-related failures also arise in other domains, and we do not test whether our findings transfer across legal subdomains, jurisdictions, alternative metrics, or other model families. Our setup further assumes access to unlabeled in-domain legal opinions for teacher distillation, an assumption that may not hold in specialized high-stakes settings. Future work should explore out-of-distribution adaptation and broader validation across datasets, metrics, and model families.

\noindent \textbf{Subjectivity of Saliency.}
Our evaluation follows the \texttt{ARC}\textsubscript{score} decomposition of arguments into atomic facts, treating all atomic facts as equally important. However, in practice, some legal arguments or supporting facts may carry greater saliency than others depending on the downstream use case. Future work should explore learning saliency weighting schemes from domain-expert feedback, although obtaining such annotations remains costly in domains such as law.

\section*{Ethical Statement}
While our approach improves argument saliency coverage, generated summaries may still omit important details or contain unsupported claims, and therefore should not be used as a substitute for professional legal analysis. Our distillation approach may also inherit biases or systematic preferences from teacher models. Finally, our experiments rely on publicly available legal documents from the \texttt{CANLII} dataset and do not involve private or personally identifiable information. Our approach is not intended for legal decision-making or autonomous legal advice.
\bibliography{custom}

@article{brown2020language,
  title={Language models are few-shot learners},
  author={Brown, Tom and Mann, Benjamin and Ryder, Nick and Subbiah, Melanie and Kaplan, Jared D and Dhariwal, Prafulla and Neelakantan, Arvind and Shyam, Pranav and Sastry, Girish and Askell, Amanda and others},
  journal={Advances in neural information processing systems},
  volume={33},
  pages={1877--1901},
  year={2020}
}

@inproceedings{elaraby2022arglegalsumm,
    title = "{A}rg{L}egal{S}umm: Improving Abstractive Summarization of Legal Documents with Argument Mining",
    author = "Elaraby, Mohamed  and
      Litman, Diane",
    editor = "Calzolari, Nicoletta  and
      Huang, Chu-Ren  and
      Kim, Hansaem  and
      Pustejovsky, James  and
      Wanner, Leo  and
      Choi, Key-Sun  and
      Ryu, Pum-Mo  and
      Chen, Hsin-Hsi  and
      Donatelli, Lucia  and
      Ji, Heng  and
      Kurohashi, Sadao  and
      Paggio, Patrizia  and
      Xue, Nianwen  and
      Kim, Seokhwan  and
      Hahm, Younggyun  and
      He, Zhong  and
      Lee, Tony Kyungil  and
      Santus, Enrico  and
      Bond, Francis  and
      Na, Seung-Hoon",
    booktitle = "Proceedings of the 29th International Conference on Computational Linguistics",
    month = oct,
    year = "2022",
    address = "Gyeongju, Republic of Korea",
    publisher = "International Committee on Computational Linguistics",
    url = "https://aclanthology.org/2022.coling-1.540",
    pages = "6187--6194",
}

@inproceedings{xu2021toward,
  title={Toward summarizing case decisions via extracting argument issues, reasons, and conclusions},
  author={Xu, Huihui and Savelka, Jaromir and Ashley, Kevin D},
  booktitle={Proceedings of the eighteenth international conference on artificial intelligence and law},
  pages={250--254},
  year={2021}
}

@inproceedings{min-etal-2023-factscore,
    title = "{FA}ct{S}core: Fine-grained Atomic Evaluation of Factual Precision in Long Form Text Generation",
    author = "Min, Sewon  and
      Krishna, Kalpesh  and
      Lyu, Xinxi  and
      Lewis, Mike  and
      Yih, Wen-tau  and
      Koh, Pang  and
      Iyyer, Mohit  and
      Zettlemoyer, Luke  and
      Hajishirzi, Hannaneh",
    editor = "Bouamor, Houda  and
      Pino, Juan  and
      Bali, Kalika",
    booktitle = "Proceedings of the 2023 Conference on Empirical Methods in Natural Language Processing",
    month = dec,
    year = "2023",
    address = "Singapore",
    publisher = "Association for Computational Linguistics",
    url = "https://aclanthology.org/2023.emnlp-main.741",
    doi = "10.18653/v1/2023.emnlp-main.741",
    pages = "12076--12100",
}

@article{hermann2015teaching,
  title={Teaching machines to read and comprehend},
  author={Hermann, Karl Moritz and Kocisky, Tomas and Grefenstette, Edward and Espeholt, Lasse and Kay, Will and Suleyman, Mustafa and Blunsom, Phil},
  journal={Advances in neural information processing systems},
  volume={28},
  year={2015}
}

@inproceedings{elaraby-etal-2023-towards,
    title = "Towards Argument-Aware Abstractive Summarization of Long Legal Opinions with Summary Reranking",
    author = "Elaraby, Mohamed  and
      Zhong, Yang  and
      Litman, Diane",
    editor = "Rogers, Anna  and
      Boyd-Graber, Jordan  and
      Okazaki, Naoaki",
    booktitle = "Findings of the Association for Computational Linguistics: ACL 2023",
    month = jul,
    year = "2023",
    address = "Toronto, Canada",
    publisher = "Association for Computational Linguistics",
    url = "https://aclanthology.org/2023.findings-acl.481",
    doi = "10.18653/v1/2023.findings-acl.481",
    pages = "7601--7612",
}

@inproceedings{elaraby-etal-2024-adding,
    title = "Adding Argumentation into Human Evaluation of Long Document Abstractive Summarization: A Case Study on Legal Opinions",
    author = "Elaraby, Mohamed  and
      Xu, Huihui  and
      Gray, Morgan  and
      Ashley, Kevin  and
      Litman, Diane",
    editor = "Balloccu, Simone  and
      Belz, Anya  and
      Huidrom, Rudali  and
      Reiter, Ehud  and
      Sedoc, Joao  and
      Thomson, Craig",
    booktitle = "Proceedings of the Fourth Workshop on Human Evaluation of NLP Systems (HumEval) @ LREC-COLING 2024",
    month = may,
    year = "2024",
    address = "Torino, Italia",
    publisher = "ELRA and ICCL",
    url = "https://aclanthology.org/2024.humeval-1.3",
    pages = "28--35",
}

@inproceedings{ravaut-etal-2024-context,
    title = "On Context Utilization in Summarization with Large Language Models",
    author = "Ravaut, Mathieu  and
      Sun, Aixin  and
      Chen, Nancy  and
      Joty, Shafiq",
    editor = "Ku, Lun-Wei  and
      Martins, Andre  and
      Srikumar, Vivek",
    booktitle = "Proceedings of the 62nd Annual Meeting of the Association for Computational Linguistics (Volume 1: Long Papers)",
    month = aug,
    year = "2024",
    address = "Bangkok, Thailand",
    publisher = "Association for Computational Linguistics",
    url = "https://aclanthology.org/2024.acl-long.153/",
    doi = "10.18653/v1/2024.acl-long.153",
    pages = "2764--2781"
}

@inproceedings{trienes2025behavioral,
    title = "Behavioral Analysis of Information Salience in Large Language Models",
    author = {Trienes, Jan  and
      Schl{\"o}tterer, J{\"o}rg  and
      Li, Junyi Jessy  and
      Seifert, Christin},
    editor = "Che, Wanxiang  and
      Nabende, Joyce  and
      Shutova, Ekaterina  and
      Pilehvar, Mohammad Taher",
    booktitle = "Findings of the Association for Computational Linguistics: ACL 2025",
    month = jul,
    year = "2025",
    address = "Vienna, Austria",
    publisher = "Association for Computational Linguistics",
    url = "https://aclanthology.org/2025.findings-acl.1204/",
    doi = "10.18653/v1/2025.findings-acl.1204",
    pages = "23428--23454",
    ISBN = "979-8-89176-256-5"
}

@inproceedings{tang-etal-2024-minicheck,
    title = "{M}ini{C}heck: Efficient Fact-Checking of {LLM}s on Grounding Documents",
    author = "Tang, Liyan  and
      Laban, Philippe  and
      Durrett, Greg",
    editor = "Al-Onaizan, Yaser  and
      Bansal, Mohit  and
      Chen, Yun-Nung",
    booktitle = "Proceedings of the 2024 Conference on Empirical Methods in Natural Language Processing",
    month = nov,
    year = "2024",
    address = "Miami, Florida, USA",
    publisher = "Association for Computational Linguistics",
    url = "https://aclanthology.org/2024.emnlp-main.499/",
    doi = "10.18653/v1/2024.emnlp-main.499",
    pages = "8818--8847"
}

@article{grattafiori2024llama,
  title={The llama 3 herd of models},
  author={Grattafiori, Aaron and Dubey, Abhimanyu and Jauhri, Abhinav and Pandey, Abhinav and Kadian, Abhishek and Al-Dahle, Ahmad and Letman, Aiesha and Mathur, Akhil and Schelten, Alan and Vaughan, Alex and others},
  journal={arXiv preprint arXiv:2407.21783},
  year={2024}
}

@inproceedings{adams-etal-2023-sparse,
    title = "From Sparse to Dense: {GPT}-4 Summarization with Chain of Density Prompting",
    author = "Adams, Griffin  and
      Fabbri, Alex  and
      Ladhak, Faisal  and
      Lehman, Eric  and
      Elhadad, No{\'e}mie",
    editor = "Dong, Yue  and
      Xiao, Wen  and
      Wang, Lu  and
      Liu, Fei  and
      Carenini, Giuseppe",
    booktitle = "Proceedings of the 4th New Frontiers in Summarization Workshop",
    month = dec,
    year = "2023",
    address = "Singapore",
    publisher = "Association for Computational Linguistics",
    url = "https://aclanthology.org/2023.newsum-1.7/",
    doi = "10.18653/v1/2023.newsum-1.7",
    pages = "68--74"
}

@inproceedings{liu-etal-2024-learning,
    title = "On Learning to Summarize with Large Language Models as References",
    author = "Liu, Yixin  and
      Shi, Kejian  and
      He, Katherine  and
      Ye, Longtian  and
      Fabbri, Alexander  and
      Liu, Pengfei  and
      Radev, Dragomir  and
      Cohan, Arman",
    editor = "Duh, Kevin  and
      Gomez, Helena  and
      Bethard, Steven",
    booktitle = "Proceedings of the 2024 Conference of the North American Chapter of the Association for Computational Linguistics: Human Language Technologies (Volume 1: Long Papers)",
    month = jun,
    year = "2024",
    address = "Mexico City, Mexico",
    publisher = "Association for Computational Linguistics",
    url = "https://aclanthology.org/2024.naacl-long.478/",
    doi = "10.18653/v1/2024.naacl-long.478",
    pages = "8647--8664"
}

@inproceedings{zhong-litman-2025-discourse,
    title = "Discourse-Driven Evaluation: Unveiling Factual Inconsistency in Long Document Summarization",
    author = "Zhong, Yang  and
      Litman, Diane",
    editor = "Chiruzzo, Luis  and
      Ritter, Alan  and
      Wang, Lu",
    booktitle = "Proceedings of the 2025 Conference of the Nations of the Americas Chapter of the Association for Computational Linguistics: Human Language Technologies (Volume 1: Long Papers)",
    month = apr,
    year = "2025",
    address = "Albuquerque, New Mexico",
    publisher = "Association for Computational Linguistics",
    url = "https://aclanthology.org/2025.naacl-long.103/",
    pages = "2050--2073",
    ISBN = "979-8-89176-189-6"
}

@inproceedings{walden-etal-2025-cross,
    title = "Cross-Document Event-Keyed Summarization",
    author = "Walden, William  and
      Kuchmiichuk, Pavlo  and
      Martin, Alexander  and
      Jin, Chihsheng  and
      Cao, Angela  and
      Sun, Claire  and
      Allen, Curisia  and
      White, Aaron Steven",
    editor = "Fei, Hao  and
      Tu, Kewei  and
      Zhang, Yuhui  and
      Hu, Xiang  and
      Han, Wenjuan  and
      Jia, Zixia  and
      Zheng, Zilong  and
      Cao, Yixin  and
      Zhang, Meishan  and
      Lu, Wei  and
      Siddharth, N.  and
      {\O}vrelid, Lilja  and
      Xue, Nianwen  and
      Zhang, Yue",
    booktitle = "Proceedings of the 1st Joint Workshop on Large Language Models and Structure Modeling (XLLM 2025)",
    month = aug,
    year = "2025",
    address = "Vienna, Austria",
    publisher = "Association for Computational Linguistics",
    url = "https://aclanthology.org/2025.xllm-1.19/",
    doi = "10.18653/v1/2025.xllm-1.19",
    pages = "218--241",
    ISBN = "979-8-89176-286-2"
}

@inproceedings{gantt-etal-2024-event,
    title = "Event-Keyed Summarization",
    author = "Gantt, William  and
      Martin, Alexander  and
      Kuchmiichuk, Pavlo  and
      White, Aaron Steven",
    editor = "Al-Onaizan, Yaser  and
      Bansal, Mohit  and
      Chen, Yun-Nung",
    booktitle = "Findings of the Association for Computational Linguistics: EMNLP 2024",
    month = nov,
    year = "2024",
    address = "Miami, Florida, USA",
    publisher = "Association for Computational Linguistics",
    url = "https://aclanthology.org/2024.findings-emnlp.431/",
    doi = "10.18653/v1/2024.findings-emnlp.431",
    pages = "7333--7345"
}

@inproceedings{elaraby-litman-2026-arc,
    title = "{ARC}: Argument Representation and Coverage Analysis for Zero-Shot Long Document Summarization with Instruction Following {LLM}s",
    author = "Elaraby, Mohamed  and
      Litman, Diane",
    editor = "Demberg, Vera  and
      Inui, Kentaro  and
      Marquez, Llu{\'i}s",
    booktitle = "Proceedings of the 19th Conference of the {E}uropean Chapter of the {A}ssociation for {C}omputational {L}inguistics (Volume 1: Long Papers)",
    month = mar,
    year = "2026",
    address = "Rabat, Morocco",
    publisher = "Association for Computational Linguistics",
    url = "https://aclanthology.org/2026.eacl-long.167/",
    doi = "10.18653/v1/2026.eacl-long.167",
    pages = "3626--3643",
    ISBN = "979-8-89176-380-7"
}

@article{ouyang2022training,
  title={Training language models to follow instructions with human feedback},
  author={Ouyang, Long and Wu, Jeffrey and Jiang, Xu and Almeida, Diogo and Wainwright, Carroll and Mishkin, Pamela and Zhang, Chong and Agarwal, Sandhini and Slama, Katarina and Ray, Alex and others},
  journal={Advances in neural information processing systems},
  volume={35},
  pages={27730--27744},
  year={2022}
}

@article{dettmers2023qlora,
  title={Qlora: Efficient finetuning of quantized llms},
  author={Dettmers, Tim and Pagnoni, Artidoro and Holtzman, Ari and Zettlemoyer, Luke},
  journal={Advances in neural information processing systems},
  volume={36},
  pages={10088--10115},
  year={2023}
}

@inproceedings{heddaya-etal-2025-casesumm,
    title = "{C}ase{S}umm: A Large-Scale Dataset for Long-Context Summarization from {U}.{S}. {S}upreme {C}ourt Opinions",
    author = "Heddaya, Mourad  and
      MacMillan, Kyle  and
      Mei, Hongyuan  and
      Tan, Chenhao  and
      Malani, Anup",
    editor = "Chiruzzo, Luis  and
      Ritter, Alan  and
      Wang, Lu",
    booktitle = "Findings of the Association for Computational Linguistics: NAACL 2025",
    month = apr,
    year = "2025",
    address = "Albuquerque, New Mexico",
    publisher = "Association for Computational Linguistics",
    url = "https://aclanthology.org/2025.findings-naacl.102/",
    doi = "10.18653/v1/2025.findings-naacl.102",
    pages = "1917--1942",
    ISBN = "979-8-89176-195-7"
}

@inproceedings{zhu2025factual,
    title = "Factual Dialogue Summarization via Learning from Large Language Models",
    author = "Zhu, Rongxin  and
      Lau, Jey Han  and
      Qi, Jianzhong",
    editor = "Rambow, Owen  and
      Wanner, Leo  and
      Apidianaki, Marianna  and
      Al-Khalifa, Hend  and
      Eugenio, Barbara Di  and
      Schockaert, Steven",
    booktitle = "Proceedings of the 31st International Conference on Computational Linguistics",
    month = jan,
    year = "2025",
    address = "Abu Dhabi, UAE",
    publisher = "Association for Computational Linguistics",
    url = "https://aclanthology.org/2025.coling-main.302/",
    pages = "4474--4492"
}

@inproceedings{narayan-etal-2018-dont,
    title = "Don{'}t Give Me the Details, Just the Summary! Topic-Aware Convolutional Neural Networks for Extreme Summarization",
    author = "Narayan, Shashi  and
      Cohen, Shay B.  and
      Lapata, Mirella",
    editor = "Riloff, Ellen  and
      Chiang, David  and
      Hockenmaier, Julia  and
      Tsujii, Jun{'}ichi",
    booktitle = "Proceedings of the 2018 Conference on Empirical Methods in Natural Language Processing",
    month = oct # "-" # nov,
    year = "2018",
    address = "Brussels, Belgium",
    publisher = "Association for Computational Linguistics",
    url = "https://aclanthology.org/D18-1206/",
    doi = "10.18653/v1/D18-1206",
    pages = "1797--1807"
}

@article{schulman2025lora,
  author = {John Schulman and Thinking Machines Lab},
  title = {LoRA Without Regret},
  journal = {Thinking Machines Lab: Connectionism},
  year = {2025},
  note = {https://thinkingmachines.ai/blog/lora/},
  doi = {10.64434/tml.20250929},
}

@inproceedings{loshchilov2019decoupled,
  title={Decoupled Weight Decay Regularization},
  author={Loshchilov, Ilya and Hutter, Frank},
  booktitle={International Conference on Learning Representations},
  year={2019}
}

@inproceedings{gao-etal-2025-train,
    title = "How to Train Long-Context Language Models (Effectively)",
    author = "Gao, Tianyu  and
      Wettig, Alexander  and
      Yen, Howard  and
      Chen, Danqi",
    editor = "Che, Wanxiang  and
      Nabende, Joyce  and
      Shutova, Ekaterina  and
      Pilehvar, Mohammad Taher",
    booktitle = "Proceedings of the 63rd Annual Meeting of the Association for Computational Linguistics (Volume 1: Long Papers)",
    month = jul,
    year = "2025",
    address = "Vienna, Austria",
    publisher = "Association for Computational Linguistics",
    url = "https://aclanthology.org/2025.acl-long.366/",
    doi = "10.18653/v1/2025.acl-long.366",
    pages = "7376--7399",
    ISBN = "979-8-89176-251-0"
}

@inproceedings{kim-rush-2016-sequence,
    title = "Sequence-Level Knowledge Distillation",
    author = "Kim, Yoon  and
      Rush, Alexander M.",
    editor = "Su, Jian  and
      Duh, Kevin  and
      Carreras, Xavier",
    booktitle = "Proceedings of the 2016 Conference on Empirical Methods in Natural Language Processing",
    month = nov,
    year = "2016",
    address = "Austin, Texas",
    publisher = "Association for Computational Linguistics",
    url = "https://aclanthology.org/D16-1139/",
    doi = "10.18653/v1/D16-1139",
    pages = "1317--1327"
}

@inproceedings{jiang-etal-2024-trisum,
    title = "{T}ri{S}um: Learning Summarization Ability from Large Language Models with Structured Rationale",
    author = "Jiang, Pengcheng  and
      Xiao, Cao  and
      Wang, Zifeng  and
      Bhatia, Parminder  and
      Sun, Jimeng  and
      Han, Jiawei",
    editor = "Duh, Kevin  and
      Gomez, Helena  and
      Bethard, Steven",
    booktitle = "Proceedings of the 2024 Conference of the North American Chapter of the Association for Computational Linguistics: Human Language Technologies (Volume 1: Long Papers)",
    month = jun,
    year = "2024",
    address = "Mexico City, Mexico",
    publisher = "Association for Computational Linguistics",
    url = "https://aclanthology.org/2024.naacl-long.154/",
    doi = "10.18653/v1/2024.naacl-long.154",
    pages = "2805--2819"
}

@inproceedings{wang-etal-2025-distilling,
  title={Distilling structured rationale from large language models to small language models for abstractive summarization},
  author={Wang, Linyong and Wu, Lianwei and Song, Shaoqi and Wang, Yaxiong and Gao, Cuiyun and Wang, Kang},
  booktitle={Proceedings of the AAAI conference on artificial intelligence},
  volume={39},
  number={24},
  pages={25389--25397},
  year={2025}
}

@inproceedings{ross2011reduction,
  title={A reduction of imitation learning and structured prediction to no-regret online learning},
  author={Ross, St{\'e}phane and Gordon, Geoffrey and Bagnell, Drew},
  booktitle={Proceedings of the fourteenth international conference on artificial intelligence and statistics},
  pages={627--635},
  year={2011},
  organization={JMLR Workshop and Conference Proceedings}
}

@inproceedings{agarwal2024policy,
  title={On-policy distillation of language models: Learning from self-generated mistakes},
  author={Agarwal, Rishabh and Vieillard, Nino and Zhou, Yongchao and Stanczyk, Piotr and Ramos Garea, Sabela and Geist, Matthieu and Bachem, Olivier},
  booktitle={International Conference on Learning Representations},
  volume={2024},
  pages={21246--21263},
  year={2024}
}

@inproceedings{hsieh-etal-2023-distilling,
    title = "Distilling Step-by-Step! Outperforming Larger Language Models with Less Training Data and Smaller Model Sizes",
    author = "Hsieh, Cheng-Yu  and
      Li, Chun-Liang  and
      Yeh, Chih-kuan  and
      Nakhost, Hootan  and
      Fujii, Yasuhisa  and
      Ratner, Alex  and
      Krishna, Ranjay  and
      Lee, Chen-Yu  and
      Pfister, Tomas",
    editor = "Rogers, Anna  and
      Boyd-Graber, Jordan  and
      Okazaki, Naoaki",
    booktitle = "Findings of the Association for Computational Linguistics: ACL 2023",
    month = jul,
    year = "2023",
    address = "Toronto, Canada",
    publisher = "Association for Computational Linguistics",
    url = "https://aclanthology.org/2023.findings-acl.507/",
    doi = "10.18653/v1/2023.findings-acl.507",
    pages = "8003--8017"
}

@article{guo-etal-2025-deepseek-r1,
  title   = {{DeepSeek-R1} Incentivizes Reasoning in {LLMs} through Reinforcement Learning},
  author  = {Guo, Daya and Yang, Dejian and Zhang, Haowei and others},
  journal = {Nature},
  volume  = {645},
  pages   = {633--638},
  year    = {2025},
  doi     = {10.1038/s41586-025-09422-z}
}

@inproceedings{muennighoff-etal-2025-s1,
    title = "s1: Simple test-time scaling",
    author = "Muennighoff, Niklas  and
      Yang, Zitong  and
      Shi, Weijia  and
      Li, Xiang Lisa  and
      Fei-Fei, Li  and
      Hajishirzi, Hannaneh  and
      Zettlemoyer, Luke  and
      Liang, Percy  and
      Cand{\`e}s, Emmanuel  and
      Hashimoto, Tatsunori",
    editor = "Christodoulopoulos, Christos  and
      Chakraborty, Tanmoy  and
      Rose, Carolyn  and
      Peng, Violet",
    booktitle = "Proceedings of the 2025 Conference on Empirical Methods in Natural Language Processing",
    month = nov,
    year = "2025",
    address = "Suzhou, China",
    publisher = "Association for Computational Linguistics",
    url = "https://aclanthology.org/2025.emnlp-main.1025/",
    doi = "10.18653/v1/2025.emnlp-main.1025",
    pages = "20275--20321",
    ISBN = "979-8-89176-332-6"
}

\appendix

\begin{figure*}[t]
\small
\centering
\begin{tikzpicture}[
  font=\small,
  node distance=18mm and 32mm,
  box/.style={
    draw=black!50,
    rounded corners=2pt,
    fill=black!3,
    align=left,
    inner sep=7pt,
    text width=6.2cm
  },
  edgelabel/.style={
    font=\small,
    fill=white,
    inner sep=1.5pt
  },
  arrow/.style={-{Stealth[length=2.2mm,width=1.8mm]}, line width=0.6pt, draw=black!60}
]

\node[box, text width=4cm] (issue) at (0,0) {\textbf{Issue:} Damage to both vehicles exceeded the insurance deductibles and both parties claim damages against each other.};

\node[box, right=of issue, text width=4cm] (conclusion) at (4,0) {\textbf{Conclusion:} Fault for this accident was attributed $10\%$ to the defendant and $90\%$ to the plaintiff.};

\node[box, below=of $(issue)!0.5!(conclusion)$, text width=5cm] (reason) {\textbf{Reason:} The plaintiful should have taken more appropriate measures to avoid the accident.};

\draw[arrow] ([xshift=10mm]issue.east) -- ([xshift=-10mm]conclusion.west)
  node[midway, above=3pt, edgelabel] {What did the court decide?};

\draw[arrow] (conclusion.south) -- (reason.north)
  node[midway, edgelabel, xshift=10mm] {Why this decision was reached?};

\end{tikzpicture}
\vspace{-1mm}
\caption{Example argument graph from a long legal opinion (CANLII): issue, conclusion, and supporting reason connected with labeled relations.}
\label{fig:canlii-arg-graph}
\end{figure*}
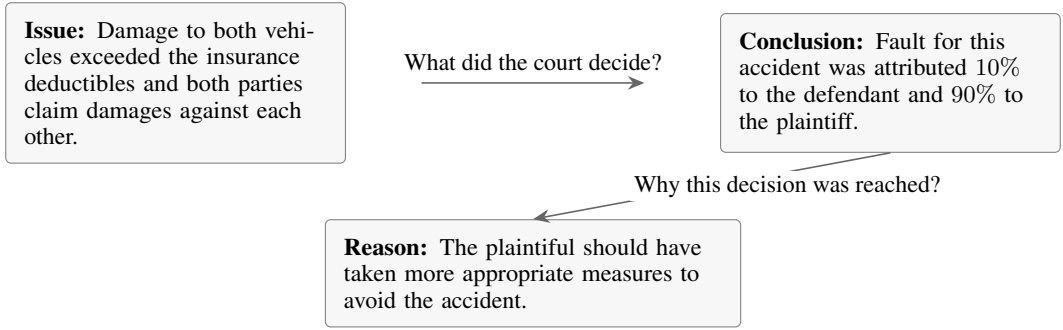
\section{Argument Roles in \texttt{CANLII}}
\label{app:canlii_examples}

Figure~\ref{fig:canlii-arg-graph} illustrates an example of argumentative roles within an annotated legal opinion. The original annotations from~\citet{xu2021toward} include only role labels, while the argumentative relations shown in the figure are added for illustration purposes.

\section{Student Training and Summarization Prompting}
\label{app:train_summ_prompt}

\subsection{Student Model Training}

\noindent \textbf{Training Setup.}
We fine-tune all student models using \texttt{QLoRA}~\cite{dettmers2023qlora} with 4-bit quantization, applying LoRA adapters to both attention and linear layers which shows comparable performance to full-finetuning across downstream applications~\cite{schulman2025lora}. We use LoRA rank $r=32$ and scaling factor $\alpha=64$, selected via grid search on \texttt{Qwen3-1.7B} over $r \in \{4,\ldots,32\}$ and $\alpha \in \{8,\ldots,64\}$. All models are trained for $3$ epochs using the \texttt{AdamW} optimizer~\cite{loshchilov2019decoupled}.

\subsection{Reasoning-Trace-Only Distillation}

In the reasoning-trace-only setting (\textbf{R}), the student is trained on teacher-generated reasoning traces without access to the final summary. To obtain these traces, we replace the standard summarization prompt (Table~\ref{tab:summarization_prompt}) with the prompt shown in Table~\ref{tab:thinking_only_prompt}.

\begin{table}[h]
\centering
\small
\begin{tabular}{p{0.9\linewidth}}
\toprule
\textit{``Think through how you would summarize the following text in \{word\_count\} words. Output only your reasoning process, and do not provide the final summary.''} \\
\bottomrule
\end{tabular}
\caption{Prompt used for reasoning-trace-only distillation.}
\label{tab:thinking_only_prompt}
\end{table}

Naively training only on reasoning traces caused degeneration at inference time: when prompted to summarize, the model often continued generating reasoning traces instead of summaries. To avoid this behavior, we retain the standard summarization prompt during training while computing the SFT loss only over reasoning-trace tokens.

\subsection{Summarization Prompt}

Table~\ref{tab:summarization_prompt} shows the prompt used for teacher summary generation and supervised fine-tuning of student models. The prompt is adapted from prior work on instruction-following summarization models~\cite{ravaut-etal-2024-context,elaraby-litman-2026-arc}.

\begin{table}[ht]
\centering
\small
\begin{tabular}{p{0.9\columnwidth}}
\toprule
\textbf{Summarization Prompt} \\
\midrule
Read the following text and summarize it: \{input\}. \\

Summarize the above text in \{length\} words. \\

\textbf{Summary:} \\
\bottomrule
\end{tabular}
\caption{Prompt used for summary generation.}
\label{tab:summarization_prompt}
\end{table}

\section{Chain-of-Arguments Prompting}
\label{app:CoA}
Table~\ref{tab:CoA_prompt} shows the prompt used for generating a structured rationale grounded in argumentative definitions. Unlike distillation, this approach avoids fine-tuning by relying on inference-time argument planning. However, it requires domain-specific argument definitions and prior task knowledge, which may not generalize across domains. Additionally, inference-time planning introduces additional token generation overhead that standard fine-tuning avoids.
\begin{table}[t]
\footnotesize
\centering
\begin{tabular}{p{0.96\columnwidth}}
\toprule
\textbf{Chain-of-Arguments (CoA) Prompt} \\
\midrule

\ttfamily

You will read a legal opinion and produce a coherent \{length\}-word summary\\
of its most salient legal arguments.\\

\\
DEFINITIONS\\

A legal argument has three components:\\
- Issue: a legal question the court addressed.\\
- Reasoning: the court's basis for its conclusion --- statutes cited,\\
precedents applied, factual findings relied on.\\
- Conclusion: how the court resolved that issue.\\

\\
PROCEDURE\\

Step 1 --- Extract. Identify every distinct argument in the opinion.\\
Order them by how decisive they are to the outcome (most decisive first).\\
Keep each component to a single line.\\

\\
Step 2 --- Summarize in \{length\} words. The summary must:\\
- cover every argument from Step 1, weighted by importance;\\
- contain only information present in the opinion --- do not introduce\\
facts, parties, statutes, holdings, or inferences from outside the\\
document;\\
- name parties, statutes, and courts exactly as the document does.\\

\\
OPINION\\

<opinion>\\
\{input\}\\
</opinion>\\

\\
OUTPUT FORMAT\\

<arguments>\\
1. Issue: ...\\
Reasoning: ...\\
Conclusion: ...\\
2. Issue: ...\\
Reasoning: ...\\
Conclusion: ...\\
</arguments>\\

<summary>\\
exactly \{length\} words\\
</summary>\\

\normalfont

\\
\bottomrule
\end{tabular}
\caption{Prompt used for the Chain-of-Arguments (CoA) baseline.}
\label{tab:CoA_prompt}
\end{table}

\section{\texttt{ARC}\textsubscript{score} Details}
\label{app:arc_score}

Given a generated summary $S$ and a set of expert-annotated salient argument roles $\mathcal{R}$, \texttt{ARC}\textsubscript{score} measures the extent to which $S$ covers the argumentative content in $\mathcal{R}$~\cite{elaraby-litman-2026-arc}. Each role $r\in\mathcal{R}$ is decomposed into atomic facts $\mathcal{F}_r$. An LLM-based verifier assigns each fact a coverage label:

\begin{equation}
\delta(f_i,S)=
\begin{cases}
1 & \text{if } f_i \text{ is correctly covered in } S,\\
0 & \text{otherwise.}
\end{cases}
\end{equation}

Role-level coverage is computed as the proportion of facts in $r$ covered by $S$:

\begin{equation}
\texttt{ARC}_{\mathrm{role}}(r,S)
=
\frac{1}{|\mathcal{F}_r|}
\sum_{f_i\in\mathcal{F}_r}
\delta(f_i,S).
\end{equation}

The final score averages role-level coverage across all annotated roles:

\begin{equation}
\texttt{ARC}\textsubscript{score}(S)
=
\frac{1}{|\mathcal{R}|}
\sum_{r\in\mathcal{R}}
\texttt{ARC}_{\mathrm{role}}(r,S).
\end{equation}

In addition to the binary coverage label, the verifier assigns an error type $e\in\{\textsc{Missing Error}, \textsc{Factual Error}\}$ for uncovered facts. \textit{Missing Errors} (\textbf{ME}) indicate salient content omitted from the summary, while \textit{Factual Errors} (\textbf{FE}) indicate contradictions or unsupported renderings of salient content. This allows us to distinguish whether improvements come primarily from better saliency coverage or reduced factual inconsistency.

\texttt{ARC}\textsubscript{score} ranges from $0$ to $1$, with higher values indicating greater coverage of salient argument roles. \citet{elaraby-litman-2026-arc} report that \texttt{ARC}\textsubscript{score} achieves the strongest correlation with expert coverage judgments among lexical, semantic, entailment-based, and decomposition-based metrics: using a \texttt{GPT-4o} verifier, it reaches Kendall's $\tau=0.465$ and Pearson's $\rho=0.593$ against expert ratings on $87$ annotated legal summaries, outperforming ROUGE ($\tau\leq0.391$), BERTScore ($\tau=0.354$), all SummaC variants ($\tau\leq0.387$), and FactScore ($\tau=0.405$). Correlations are further improved when an open-weight \texttt{DeepSeek-R1-Distill-Qwen-14B} verifier is used ($\tau=0.509$, $\rho=0.638$).

\section{Scaling Training Context Length}
\label{app:len_scale}

\begin{figure}[ht]
    \centering
    \includegraphics[width=.95\columnwidth]{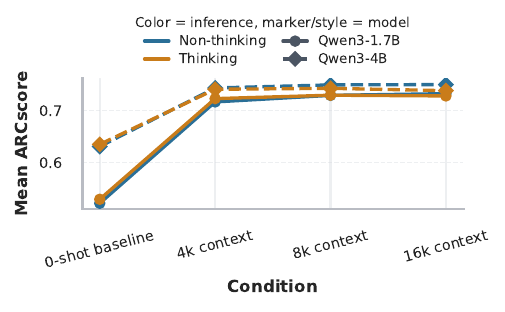}
    \caption{
    Context-length distillation across student models and inference strategies  
    }
    \label{fig:context_len_distillation_by_model}
\end{figure}

We investigate whether exposing student models to longer training documents improves argument saliency coverage. Using the same $1{,}000$ training documents, we resample inputs to include contexts up to $8$k and $16$k tokens while maintaining a balanced length distribution across groups using $2$k-token bins. We evaluate summary-only distillation from \texttt{Qwen3-14B} on two representative student sizes: \texttt{Qwen3-1.7B} and \texttt{Qwen3-4B}. Figure~\ref{fig:context_len_distillation_by_model} shows that the dominant improvement comes from distillation itself, with the jump from zero-shot to $4$k training contexts substantially outweighing gains from further context scaling. Beyond $4$k, longer training contexts provide only marginal additional improvements, particularly in non-thinking inference mode in the $4B$ student model. These findings suggest that while longer contexts can modestly improve argument coverage, their effect is secondary to distillation itself, consistent with our broader data-efficiency results.

\section{Significance Testing}
\label{app:significance}

To complement the Mann--Whitney U tests reported in
Table~\ref{tab:arc_1000_distillation}, we conduct paired Wilcoxon
signed-rank tests over the same test documents, comparing each distillation
setting against the zero-shot baseline (\textbf{B}), expert tuning
(\textbf{E}), and the Chain-of-Arguments baseline (\textbf{CoA}).
Table~\ref{tab:significance} reports $1{,}000$-example summary distillation
(non-thinking inference); each cell gives the comparison baseline's mean,
the difference, and the Wilcoxon $p$-value.

Distillation improves significantly over the zero-shot baseline for every
student under both teachers (all $p<0.001$). Against expert tuning,
distillation from the stronger \texttt{Qwen3-14B} teacher is significantly
better for \texttt{Qwen3-1.7B} and larger, and statistically comparable at
\texttt{Qwen3-0.6B} ($+0.022$, $p=0.09$); with the weaker \texttt{GPT-5-mini}
teacher, distillation exceeds expert tuning from \texttt{Qwen3-1.7B} upward
but is slightly below it at \texttt{Qwen3-0.6B} ($-0.046$, $p=0.01$). A
similar pattern holds against \textbf{CoA}: the \texttt{Qwen3-14B} teacher is
significantly better for all students, while \texttt{GPT-5-mini} exceeds
\textbf{CoA} for the larger students but not for \texttt{Qwen3-0.6B}, where
\textbf{CoA} remains competitive.

\begin{table*}[t]
\small
\centering
\begin{tabular}{l l c ccc}
\toprule
\textbf{Student} & \textbf{Teacher} & \textbf{ARC}
  & \textbf{vs B} & \textbf{vs E} & \textbf{vs CoA} \\
  & & & ($\Delta$/$p$) & ($\Delta$/$p$) & ($\Delta$/$p$) \\
\midrule
\multirow{2}{*}{Qwen3-0.6B}
  & Qwen3-14B  & .621 & $+.213$/${<}.001$ & $+.022$/$.09$  & $+.044$/${<}.001$ \\
  & GPT-5-mini & .553 & $+.145$/${<}.001$ & $-.046$/$.011$ & $-.025$/$.020$ \\
\addlinespace
\multirow{2}{*}{Qwen3-1.7B}
  & Qwen3-14B  & .717 & $+.195$/${<}.001$ & $+.118$/${<}.001$ & $+.093$/${<}.001$ \\
  & GPT-5-mini & .651 & $+.129$/${<}.001$ & $+.052$/$.002$    & $+.027$/$.010$ \\
\addlinespace
\multirow{2}{*}{Qwen3-4B}
  & Qwen3-14B  & .744 & $+.112$/${<}.001$ & $+.173$/${<}.001$ & $+.065$/${<}.001$ \\
  & GPT-5-mini & .712 & $+.081$/${<}.001$ & $+.141$/${<}.001$ & $+.034$/${<}.001$ \\
\addlinespace
\multirow{2}{*}{Qwen3-8B}
  & Qwen3-14B  & .781 & $+.123$/${<}.001$ & $+.255$/${<}.001$ & $+.083$/${<}.001$ \\
  & GPT-5-mini & .741 & $+.083$/${<}.001$ & $+.215$/${<}.001$ & $+.043$/${<}.001$ \\
\bottomrule
\end{tabular}
\caption{
Paired Wilcoxon signed-rank tests for $1{,}000$-example summary distillation
(non-thinking inference). Each cell reports the difference ($\Delta$) in
\texttt{ARC}\textsubscript{score} and the $p$-value against the zero-shot
baseline (\textbf{B}), expert tuning (\textbf{E}), and Chain-of-Arguments
(\textbf{CoA}). Baseline means: \textbf{B} $=.408/.522/.631/.658$;
\textbf{E} $=.599/.599/.571/.526$; \textbf{CoA} $=.578/.624/.679/.698$
for the four student sizes.
}
\label{tab:significance}
\end{table*}

\section{Perplexity Analysis: Explaining Teacher Impact}
\label{app:perplexity}

\begin{figure*}[ht]
    \centering
    \includegraphics[width=0.9\textwidth]{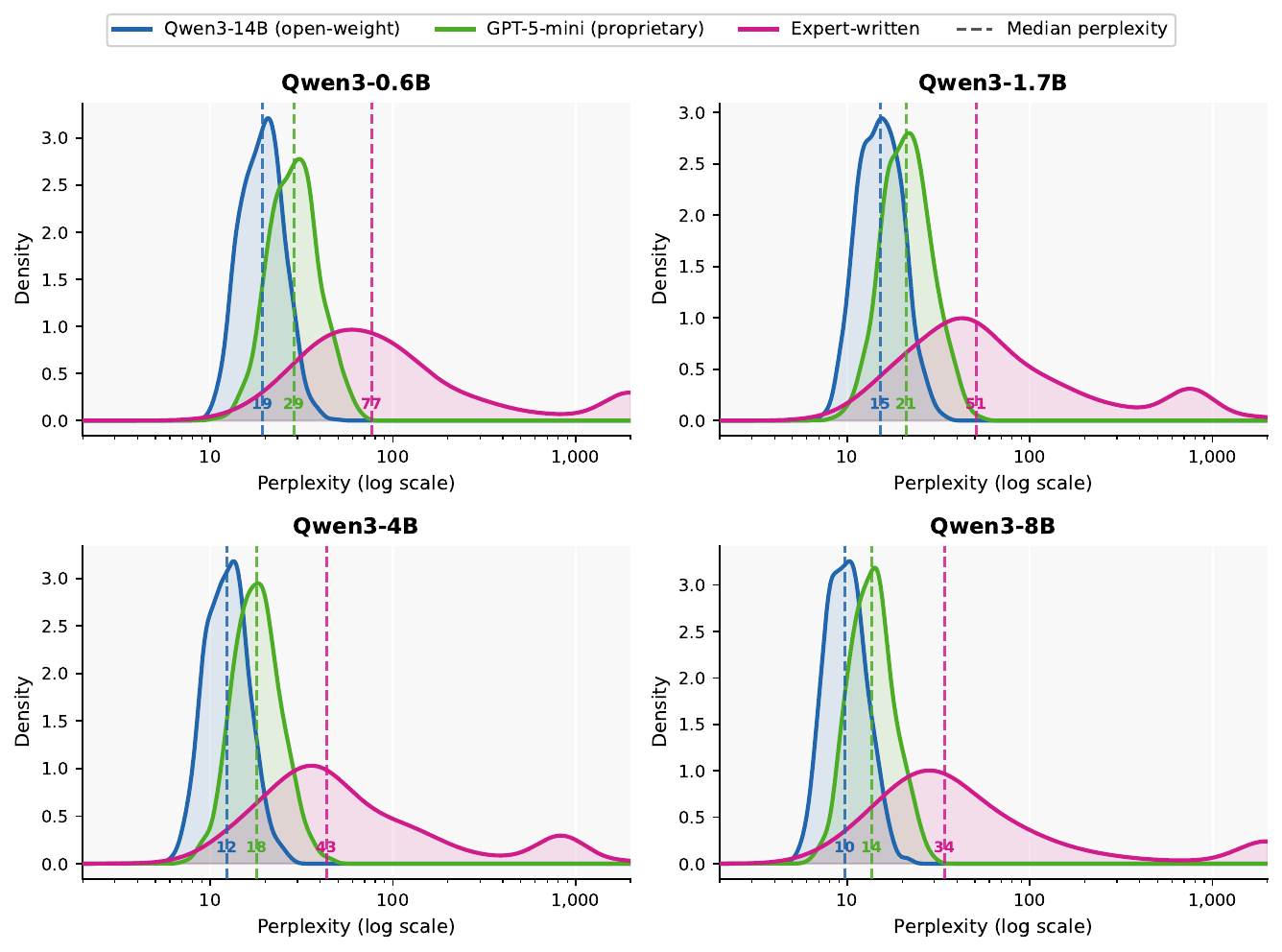}
    \caption{
    Perplexity distributions of teacher-generated and expert-written 
    summaries under each base student model (before fine-tuning). 
    Dashed lines indicate medians. 
    }
    \label{fig:ppl_dist}
\end{figure*}

To investigate why \texttt{Qwen3-14B} summaries yield stronger 
distillation signal than \textit{GPT-5-mini} or expert-written 
summaries, we analyze the perplexity of each summary type under 
the base student models prior to fine-tuning. Since autoregressive 
language models are trained to minimize the negative log-likelihood 
of their training data~\cite{brown2020language}, perplexity directly 
reflects how compatible a given text is with the student's learned 
distribution. Summaries with lower 
perplexity under the student are closer to its pretraining 
distribution, providing a more on-policy supervision 
signal~\cite{agarwal2024policy} and reducing the 
distribution mismatch that standard off-policy fine-tuning must 
overcome~\cite{ross2011reduction}.

Figure~\ref{fig:ppl_dist} shows that \texttt{Qwen3-14B} summaries 
consistently yield the lowest median perplexity across all student 
sizes (median: $9.7$--$19.3$), followed by \textit{GPT-5-mini} 
(median: $13.6$--$28.8$), with expert-written summaries exhibiting 
substantially higher and more variable perplexity (median: 
$34.1$--$76.5$, with a heavy tail extending to thousands). This 
ordering is consistent across all four student model sizes and 
directly mirrors the distillation performance ranking in 
Table~\ref{tab:arc_1000_distillation}: the teacher whose summaries 
are most compatible with the student's distribution produces the 
strongest coverage gains. Notably, as student model size increases, 
perplexity under \texttt{Qwen3-14B} summaries decreases (from $19.3$ 
at 0.6B to $9.7$ at 8B), suggesting that larger students within the 
same model family are increasingly well-aligned with the teacher's 
output distribution.
This distributional alignment may partly explain why expert fine-tuning degrades most severely at 8B, where the gap between expert and teacher outputs is largest relative to the student’s strong zero-shot prior.

We emphasize that this analysis is correlational and offered as a partial
diagnostic of distributional alignment between teachers and students,
not as a complete causal account of teacher impact. A rigorous
explanation would additionally need to control for teacher quality on
the summarization task itself and for surface-level features such as
lexical and semantic similarity between teacher outputs and expert
references. We leave a fuller analysis spanning multiple student--teacher
pairs along these axes to future work.

\section{Analysis of Expert-Tuning Underperformance}
\label{app:expert_tuning}

To better understand why fine-tuning on expert-written summaries (\textbf{E})
underperforms teacher distillation, we conduct two additional analyses.

\noindent \textbf{Controlling for the student model family.}
A natural concern is that distillation benefits simply because the
open-weight teacher (\texttt{Qwen3-14B}) shares an architecture and
tokenizer with the \texttt{Qwen3} students. To test this, we tune two
additional students from different families---\texttt{Qwen2.5-3B} and
\texttt{Llama-3.2-3B}---on \texttt{Qwen3-14B} summaries versus expert
summaries. The pattern is consistent with our main results: expert tuning
reduces \texttt{ARC}\textsubscript{score} below the zero-shot baseline for
both \texttt{Qwen2.5-3B} ($0.680\rightarrow0.241$) and \texttt{Llama-3.2-3B}
($0.691\rightarrow0.482$), whereas \texttt{Qwen3-14B} distillation improves
them to $0.767$ and $0.746$, respectively. This indicates that the gains
arise from the teacher-generated targets rather than merely from
fine-tuning on in-domain documents or from shared model family.

\noindent \textbf{Failure modes of expert tuning.}
Inspecting expert-tuned generations reveals scale-dependent failure modes.
For smaller students, low-scoring outputs often collapse into a single
sentence---for example, a \texttt{Qwen3-0.6B} summary states only that the
plaintiff was awarded a fixed sum in costs while omitting the underlying
dispute and legal holding entirely (\texttt{ARC}\textsubscript{score} $0$).
Larger students generally remain fluent and produce longer summaries, but
may focus on procedural background or one party's allegations while missing
the court's decision; one $265$-word \texttt{Qwen3-4B} summary received an
\texttt{ARC}\textsubscript{score} of $0$ despite being well formed. A
minority of larger-model outputs also exhibit repetitive looping, restating
the same proposition until truncation. These failures are consistent with
teacher-generated summaries providing more learnable supervision under our
training setup, and with the high expert-target perplexity reported in
Appendix~\ref{app:perplexity}, suggesting that $1{,}000$ examples are
insufficient for stable adaptation to the expert distribution.

\section{Relative Change in Factual and Missing Errors}
\label{app:relative_reduction}

\definecolor{teacherQwen}{HTML}{534AB7}
\definecolor{teacherGPT}{HTML}{BA7517}
\newcommand{\cbox}[1]{\textcolor{#1}{\rule{0.7em}{0.7em}}}

\begin{figure}[ht]
    \centering
    \includegraphics[width=1.\columnwidth]{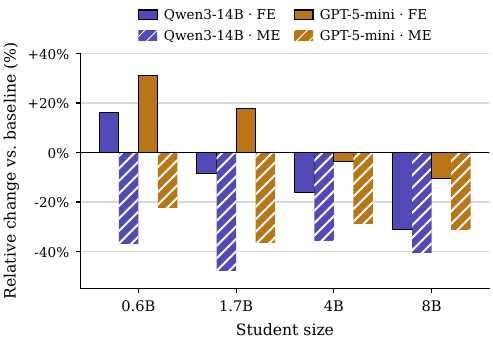}
    \caption{
    Relative change in Factual Errors (FE, solid) and Missing Errors (ME, hatched) under summary-only distillation relative to the zero-shot baseline, grouped by student size and teacher (\cbox{teacherQwen}~Qwen3-14B, \cbox{teacherGPT}~GPT-5-mini). Results are shown under thinking-mode inference; negative values indicate fewer errors.
    }
\label{fig:relative_reduction}
\end{figure}

Figure~\ref{fig:relative_reduction} reveals two patterns.
Coverage gains are immediate: ME drops at every student size for both
teachers, with the 1.7B student showing the largest relative reduction
($-48\%$ with Qwen3-14B).
Factuality gains, in contrast, require a capacity threshold:
FE \emph{increases} at 0.6B for both teachers and only begins decreasing
once the student is large enough to absorb the factuality signal --
between 0.6B and 1.7B for Qwen3-14B, and between 1.7B and 4B for GPT-5-mini.
Across every size, Qwen3-14B dominates GPT-5-mini on both metrics, with
the FE gap widening as the student grows.

Table~\ref{tab:ezurike-case-study} shows the atomic facts decomposed from salient
rhetorical roles and the per-fact verifier verdicts for summaries generated
by the baseline and the two fine-tuned models. The \texttt{ARC}\textsubscript{score}
gains over the baseline come primarily from a reduction in Missing Errors
(\textbf{ME}: $10 \rightarrow 5 \rightarrow 3$), while Factual Errors remain
low and roughly constant (\textbf{FE}: $1 \rightarrow 2 \rightarrow 1$).

\providecommand{\factYes}{\ding{51}}
\providecommand{\factNo}{{\color{black!45}\textendash}}
\definecolor{factHL}{HTML}{FFF1A8}
\providecommand{\sethlcolor}[1]{}      
\sethlcolor{factHL}
\providecommand{\hl}[1]{\textcolor{factHL!35!black}{\textbf{#1}}}
\providecommand{\fref}[1]{\textsuperscript{\textcolor{black!55}{\,#1}}}

\providecommand{\factYes}{\ding{51}}
\providecommand{\factMiss}{{\color{black!45}\textendash}}
\providecommand{\factErr}{{\color{red!75!black}\ding{55}}}
\definecolor{factHL}{HTML}{FFF1A8}
\providecommand{\sethlcolor}[1]{}      
\sethlcolor{factHL}
\providecommand{\hl}[1]{\textcolor{factHL!35!black}{\textbf{#1}}}
\providecommand{\fref}[1]{\textsuperscript{\textcolor{black!55}{\,#1}}}

\begin{table*}[t]
\centering
\small
\setlength{\tabcolsep}{4pt}

\renewcommand{\arraystretch}{1.05}
\begin{tabular}{@{}r @{\hspace{6pt}}
                >{\RaggedRight\arraybackslash}p{0.78\linewidth}
                @{\hspace{10pt}}c@{\hspace{12pt}}c@{\hspace{12pt}}c@{}}
\toprule
\textbf{\#} & \textbf{Atomic fact (rhetorical role)} &
\textbf{(a)} & \textbf{(b)} & \textbf{(c)} \\
\midrule
1 & Custodial arrangements for the parties' two youngest children were at issue. \hfill\mbox{\textit{(Issue)}}
  & \factMiss & \factMiss & \factMiss \\
2 & The two youngest children are aged 15 and 13. \hfill\mbox{\textit{(Issue)}}
  & \factMiss & \factErr  & \factMiss \\
3 & The two youngest children currently lived with the father the majority of the time. \hfill\mbox{\textit{(Issue)}}
  & \factMiss & \factErr  & \factMiss \\
\addlinespace[2pt]
4 & Children are placed in the custody of the mother. \hfill\mbox{\textit{(Conclusion)}}
  & \factYes  & \factYes  & \factYes  \\
5 & Children are placed in the primary care of the mother. \hfill\mbox{\textit{(Conclusion)}}
  & \factYes  & \factYes  & \factYes  \\
\addlinespace[2pt]
6 & Until the separation, the mother had been the primary caretaker of the children. \hfill\mbox{\textit{(Reason 1)}}
  & \factMiss & \factYes  & \factYes  \\
7 & The mother appeared to have a stronger emotional bond with the children. \hfill\mbox{\textit{(Reason 1)}}
  & \factMiss & \factYes  & \factYes  \\
8 & The father had been physically, emotionally, and verbally abusive to the wife throughout the marriage. \hfill\mbox{\textit{(Reason 1)}}
  & \factMiss & \factMiss & \factErr  \\
9 & The father seemed incapable of putting the children's needs above his own. \hfill\mbox{\textit{(Reason 1)}}
  & \factMiss & \factYes  & \factYes  \\
10 & The mother was much more likely than the father to encourage the children to have contact with the other parent. \hfill\mbox{\textit{(Reason 1)}}
  & \factMiss & \factYes  & \factYes  \\
\addlinespace[2pt]
11 & The father was retired. \hfill\mbox{\textit{(Reason 2)}}
  & \factMiss & \factMiss & \factYes  \\
12 & The father might have more time to spend with the children. \hfill\mbox{\textit{(Reason 2)}}
  & \factErr  & \factMiss & \factYes  \\
13 & Having more time to spend with the children did not outweigh all of the other factors. \hfill\mbox{\textit{(Reason 2)}}
  & \factMiss & \factMiss & \factYes  \\
\midrule
\multicolumn{2}{r@{\hspace{10pt}}}{\textit{\factYes\ Supported (of 13)}}
  & 2 & 6 & \textbf{9} \\
  \multicolumn{2}{r@{\hspace{10pt}}}{\textit{\factMiss\ Missing (of 13)}}
  & 10 & 5 & \textbf{3} \\
\multicolumn{2}{r@{\hspace{10pt}}}{\textit{\factErr\ Contradicted (of 13)}}
  & 1 & 2 & 1 \\
\bottomrule
\end{tabular}

\vspace{4pt}

\renewcommand{\arraystretch}{1.15}
\footnotesize
\setlength{\tabcolsep}{6pt}
\begin{tabular}{@{}
  >{\RaggedRight\arraybackslash}p{0.305\linewidth}
  >{\RaggedRight\arraybackslash}p{0.305\linewidth}
  >{\RaggedRight\arraybackslash}p{0.305\linewidth}@{}}
\toprule
\textbf{(a) Qwen3-0.6B (baseline)} &
\textbf{(b) + GPT-5-mini teacher} &
\textbf{(c) + Qwen3-14B teacher} \\
\textit{302 words} &
\textit{321 words} &
\textit{327 words} \\
\midrule

The case between Christie Chika Ezurike and Levi Chucks Ezurike, which resulted in a divorce, was brought to the Supreme Court of Nova Scotia (Family Division) in 2006.
[\ldots] The parents were unable to agree on custody arrangements or the division of assets or child support.
[\ldots] The marriage lasted over 20 years, and the Respondent worked in the federal government and maintained a taxi driver's license.
[\ldots]
\hl{The court ruled that the Petitioner should have custody of the children and primary care}\fref{4,\,5}, despite the Respondent's financial difficulties.
The Petitioner emphasized her commitment to the children's well-being, while the Respondent was unable to provide adequate care.
The case highlighted the importance of the children's best interests [\ldots] and the need for the children to have a positive relationship with both parents.
&

This is a divorce judgment from the Nova Scotia Family Division, issued by Justice Leslie J.\ Dellapinna.
Christie Ezurike, the Petitioner, sought divorce and custody of her two older children [\ldots]; the Respondent, Levi Ezurike, opposed [her] application and sought custody of Nneka and access to Abuchi. [\ldots]
\hl{The Petitioner testified she is the primary caregiver to both children and that she has made effort to provide for them}\fref{5,\,6}, though the Respondent contested that effort.
[\ldots]
\hl{Christie also expressed concern about the Respondent's treatment of the children and his belief that they should not be left unsupervised for extended periods.}\fref{7,\,9,\,10}
[\ldots]
\hl{The court concluded Christie has the better interests of the children and granted her divorce application, with custody of Abuchi and Nneka determined by the trial judge.}\fref{4}
&

In \emph{Ezurike v.\ Ezurike}, the Supreme Court of Nova Scotia ruled on custody and access arrangements for four children of the divorcing couple. [\ldots] \hl{The Respondent, now retired from federal service}\fref{11}, resides in a reconstructed home with the children [\ldots]. \hl{He has spent considerable time with the children}\fref{12} [\ldots]. \hl{The Petitioner, a successful social worker, has focused on providing emotional and practical care for the children}\fref{6,\,7} [\ldots].

[\ldots] \hl{The Respondent's emotional detachment and lack of understanding of the children's challenges}\fref{9,\,10} [\ldots] raised concerns.

Ultimately, \hl{the court concluded that the Petitioner should have custody and primary care of the children}\fref{4,\,5}, \hl{with the Respondent sharing reasonable access on alternate weekends}\fref{13}. [\ldots]
\\

\midrule
\multicolumn{1}{@{}c}{\textbf{ARCScore: 0.250}} &
\multicolumn{1}{c}{\textbf{ARCScore: 0.450}} &
\multicolumn{1}{c@{}}{\textbf{ARCScore: 0.700}} \\
\bottomrule
\end{tabular}
\caption{\textbf{Legal opinion: \emph{Ezurike v.\ Ezurike} (\texttt{d\_2006nssc73.txt}).}
Per-fact coverage (top) and generated summaries (bottom) for three systems: non-tuned Qwen3-0.6B~(a), and the same model tuned with a GPT-5-mini teacher~(b) or a Qwen3-14B teacher~(c).
Each cell shows the verifier's verdict: \factYes\ supported, \factMiss\ missing (fact absent from the summary), \factErr\ contradicted (summary asserts something inconsistent with the fact).
\hl{Highlighted} spans were accepted by the verifier as supporting an atomic fact; the superscript indicates which fact (e.g.\ \fref{4}).
\texttt{ARC}\textsubscript{score} is the mean recall across the four rhetorical components; passages of the summaries not bearing on any atomic fact have been elided as \mbox{[\ldots]}.}
\label{tab:ezurike-case-study}
\end{table*}

\section{Impact of LoRA Rank on Distillation Performance}
\label{app:lora_impact}

\begin{figure}[ht]
    \centering
    \includegraphics[width=1.0\columnwidth]{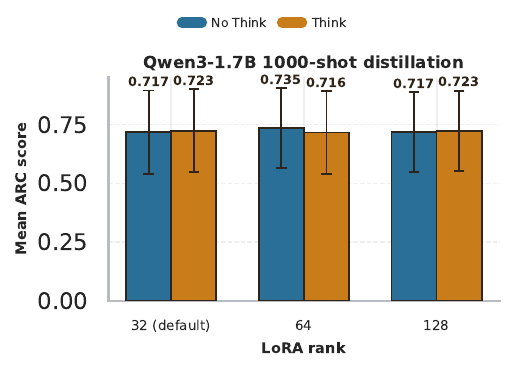}
    \caption{
    Impact of increasing LoRA rank on \texttt{Qwen3-1.7B} student.  
    }
    \label{fig:lora_impact}
\end{figure}

To verify that the observed saturation in distillation performance is not 
an artifact of limited model capacity, we increase the LoRA rank beyond 
the default $r=32$ to $r \in \{64, 128\}$, where $r=128$ approaches 
the capacity of full-parameter tuning~\cite{schulman2025lora}. 
Figure~\ref{fig:lora_impact} shows that increasing the rank yields no 
significant improvement ($p > 0.05$ in all cases), and in most settings 
performance remains flat or slightly degrades when thinking mode is 
enabled at inference. These results confirm that the saturation pattern 
reported in ~\ref{subsec:scale_samples} reflects a genuine property of the 
distillation signal rather than a capacity bottleneck in the student model. 

\section{Few-shot Distillation by Student Model}
\label{app:few_shot_detailed}

\begin{figure*}[ht]
    \centering
    \includegraphics[width=1.\textwidth]{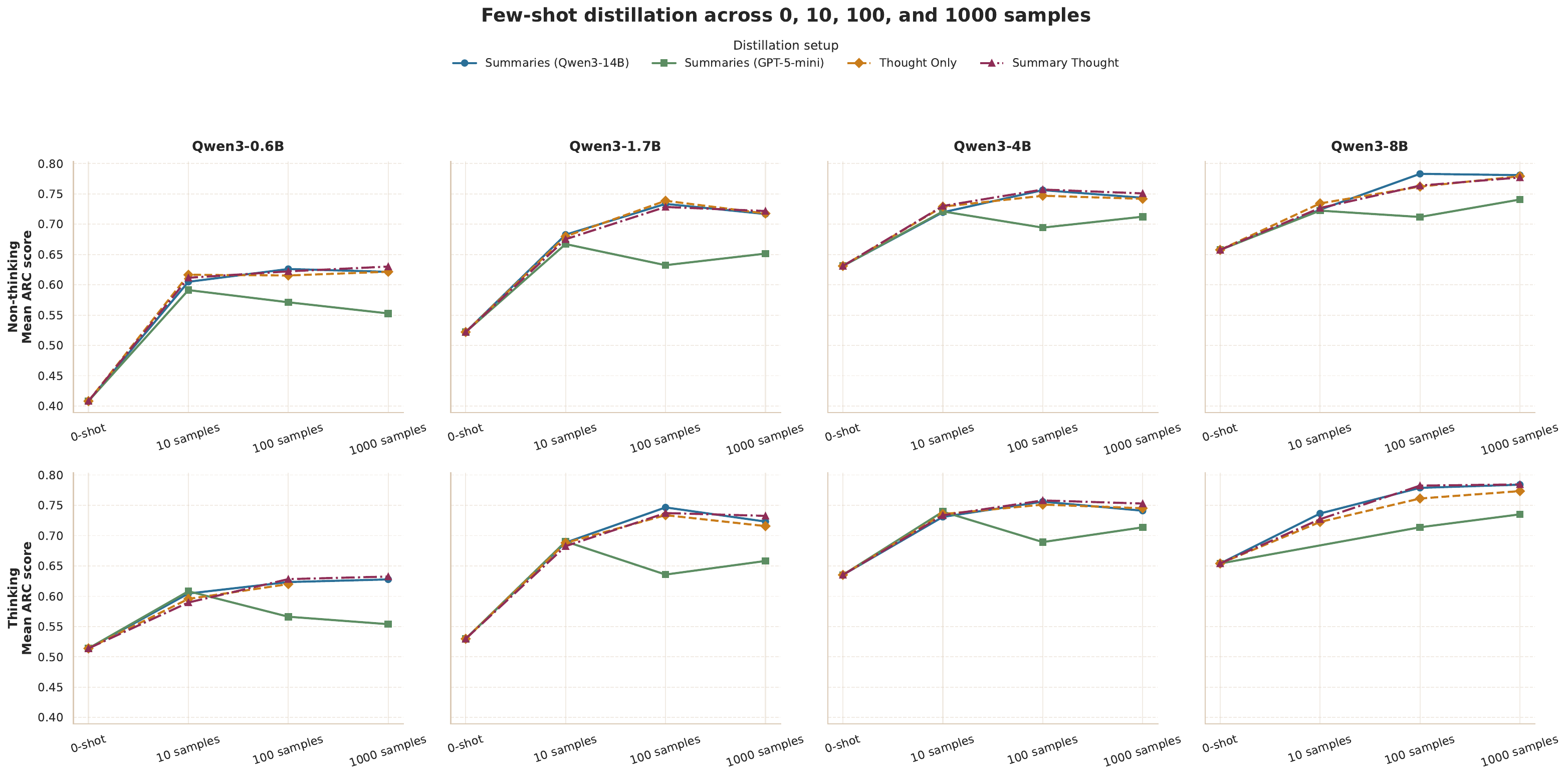}
    \caption{
    Few-shot distillation across models and inference strategies  
    }
    \label{fig:fewshot_distillation_breakdown}
\end{figure*}

Figure~\ref{fig:fewshot_distillation_breakdown} disaggregates few-shot distillation by student model size and inference mode. The main trend is consistent with Figure~\ref{fig:fewshot_distillation}: most of the improvement occurs with only $10$ training documents, and scaling to $100$ or $1{,}000$ documents provides diminishing and sometimes inconsistent gains. This pattern holds across student sizes, suggesting that argument-saliency distillation is highly data-efficient rather than primarily data-scale driven. Moreover, reasoning-chain-only (\textbf{R}) and summary+reasoning-chain (\textbf{SR}) supervision closely follow summary-only distillation from \texttt{Qwen3-14B}, suggesting that reasoning traces do not provide a consistently stronger supervision signal than summaries alone. Finally, \textit{GPT-5-mini} exhibits a distinct instability, with performance often decreasing at $100$ documents before recovering at $1{,}000$, although this effect is weaker for the \texttt{Qwen3-8B} student.

\noindent \textbf{Fewshot sampling sensitivity to randomization}
To confirm that the few-shot trend is not sensitive to the particular
documents sampled, Table~\ref{tab:seed_robustness} reports the $10$- and
$100$-document settings under two random seeds for \texttt{Qwen3-1.7B} and
\texttt{Qwen3-4B} (summary-only distillation from \texttt{Qwen3-14B},
non-thinking inference). Across both seeds, most of the gain over the
zero-shot baseline is already realized at $10$ documents and saturates by
$100$; the two seeds agree closely, differing by at most $0.031$
\texttt{ARC}\textsubscript{score} at the $10$-document setting and by
$\leq0.015$ at $100$ documents.

\begin{table}[ht]
\small
\centering
\begin{adjustbox}{max width=\columnwidth}
\begin{tabular}{l l cccc}
\toprule
\textbf{Model} & \textbf{Seed}
  & \textbf{B ($0$)} & \textbf{$10$} & \textbf{$100$} & \textbf{$1000$} \\
\midrule
\multirow{2}{*}{Qwen3-1.7B}
  & seed 42  & .522 & .683 & .734 & .717 \\
  & seed 123 & .522 & .652 & .749 & .717 \\
\addlinespace
\multirow{2}{*}{Qwen3-4B}
  & seed 42  & .631 & .720 & .756 & .744 \\
  & seed 123 & .631 & .735 & .757 & .744 \\
\bottomrule
\end{tabular}
\end{adjustbox}
\caption{
Few-shot distillation across two random seeds (summary-only distillation
from \texttt{Qwen3-14B}, non-thinking inference). The trend is consistent
across seeds: gains are largely realized by $10$ documents and saturate by
$100$.
}
\label{tab:seed_robustness}
\end{table}

\end{document}